\documentclass{ifacconf}

\usepackage{graphicx}      
\usepackage{natbib}        
\usepackage{amsmath}
\usepackage{subfigure}
\usepackage{enumitem}
\usepackage{mathtools}
\begin{document}
\begin{frontmatter}

\title{Design of Reward Function on Reinforcement Learning for Automated Driving} 

\author[HG]{Takeru Goto} 
\author[HM]{Yuki Kizumi}
\author[HM]{Shun Iwasaki}

\address[HG]{Innovative Research Excellence, Honda R\&D Co., Ltd., Tokyo, Japan (e-mail: takeru\_goto@jp.honda).}
\address[HM]{Business Development Operations, Honda Motor Co., Ltd., Tokyo, Japan.}

\begin{abstract}                
This paper proposes a design scheme of reward function that constantly evaluates both driving states and actions for applying reinforcement learning to automated driving. In the field of reinforcement learning, reward functions often evaluate whether the goal is achieved by assigning values such as +1 for success and -1 for failure. This type of reward function can potentially obtain a policy that achieves the goal, but the process by which the goal is reached is not evaluated. However, process to reach a destination is important for automated driving, such as keeping velocity, avoiding risk, retaining distance from other cars, keeping comfortable for passengers. Therefore, the reward function designed by the proposed scheme is suited for automated driving by evaluating driving process. The effects of the proposed scheme are demonstrated on simulated circuit driving and highway cruising. Asynchronous Advantage Actor-Critic is used, and models are trained under some situations for generalization. The result shows that appropriate driving positions are obtained, such as traveling on the inside of corners, and rapid deceleration to turn along sharp curves. In highway cruising, the ego vehicle becomes able to change lane in an environment where there are other vehicles with suitable deceleration to avoid catching up to a front vehicle, and acceleration so that a rear vehicle does not catch up to the ego vehicle.
\end{abstract}


\end{frontmatter}

\section{Introduction}
Reinforcement learning(RL) is a kind of machine learning that obtains an action policy which enables an agent to acquire greater rewards. \cite{mnih2013playing,mnih2015human} showed that deep reinforcement learning(DRL) which combines a deep neural network and reinforcement learning enables to obtain scores exceeding those of humans in some Atari games. This has promoted researches on DRL.

Meanwhile, appropriate reward functions are needed to generate appropriate policies. The game “Pong” is often used to evaluate RL algorithms, and the reward function commonly used in this case assigns a score of +1 when the opponent cannot return the ball, -1 when the actor cannot return the ball, and 0 for other states, e.g. \cite{mnih2015human}. This paper refers to this type of reward function as a purpose-oriented reward function. Purpose-oriented reward functions are thought to be suited to tasks where the goal is clear and any ways of achieving that goal are acceptable. This means that goals may be achieved by actions that are not easily imaginable by humans.

However, in case of automated driving, it is not enough to simply arrive at the destination without accident. The vehicle needs to drive over the assigned route while maintaining an appropriate distance between vehicles, not coming too close to pedestrians, obeying traffic rules such as speed limits, and taking into account ride comfort. It is a challenge to obtain an appropriate policy for these types of tasks using +1/-1 type purpose-oriented rewards. This calls for a reward function that constantly evaluates the appropriateness of states and actions. This paper refers to this type of reward function as a process-oriented reward function. However, process-oriented reward functions have a high degree of freedom and it is a challenge to design a reward function that can obtain an appropriate policy.

There are a lot of works for applying RL to automated driving. As an early study, \cite{stafylopatis1998autonomous} describes that the appropriate heading direction for avoiding walls can be learned with RL. \cite{michels2005high} realized obstacle avoidance. Recently, \cite{schester2021automated} proved that multi-agent DRL is effective for highway lamp merge. In these works, there are some types of rewards. The penalty on other objects or static obstacles is often used, e.g. \cite{stafylopatis1998autonomous,michels2005high}. \cite{desjardins2011cooperative} adopts the distance from obstacles or center of lane as the similar reason. That can also consider the appropriate inter-vehicle distance and position in lane. In works where RL was applied to racing simulators, e.g. \cite{lillicrap2015continuous, perot2017end}, reward functions were calculated from ego velocity and offset from the center of a road. As another point of view, drivers’ comfortability is important for automated driving. \cite{ye2020automated} considers the difference between actual and desired velocity, acceleration and jerk for that objective. These works use process-oriented reward functions partially, but each reward function is designed for each task.

Therefore, this paper proposes a general design scheme of process-oriented reward function for automated driving. This scheme calculates an evaluation value between 0 and 1 for all evaluation items, and determines the reward based on the product of those values. A separate evaluation function can be applied to each evaluation item, which facilitates a process-oriented reward function design. In addition, the proposed scheme is applied to circuit driving and highway cruising to investigate application to automated driving.

\section{Previous Research}
\subsection{Asynchronous Advantage Actor-Critic}
This paper uses Asynchronous Advantage Actor-Critic (A3C) developed by \cite{mnih2016asynchronous} as the RL algorithm. A3C is an Actor-Critic type algorithm, and can output continuous action. In addition, multiple agents simultaneously learn asynchronously, so this can utilize computation resources efficiently.

A3C has a policy function $\pi(a_t|s_t;\theta)$ and an estimated state value function $V(s_t;\theta_v)$. $a_t$ and $s_t$ are the respective action and state at time $t$, $\theta$ and $\theta_v$ are parameters that respectively determine the policy function and state value function. The action is a random variable that conforms to the policy function, and the probability of each action being selected is determined when the state is determined. The state value function is the expected value of the future return $R$ in state $s_t$. Return $R$ is the discounted sum of the rewards from time $t$ onward, and is as follows.
\begin{equation}
R=\sum_{i=0}^\infty\gamma^ir_{t+i}
\label{eq:income}
\end{equation}
$\gamma$ is the discount factor, and $r_t$ is the reward given at time $t$.
The policy function and state function are updated when an action has been performed $t_{max}$ times or when the terminal state is reached. The policy function update amount is determined  as $\nabla_{\theta'}\log\pi(a_t|s_t;\theta')A(s_t,a_t;\theta,\theta_v)$. The update amout of the state function is determined as ${\partial}A(s_t,a_t;\theta,\theta'_v)/\partial\theta'_v$. $A(s_t,a_t;\theta,\theta_v)$ is called the advantage function as follows.
\begin{equation}
A(s_t,a_t;\theta,\theta_v)\!=\!\sum_{i=0}^{k-1}\gamma^ir_{t\!+\!i}\!+\!\gamma^kV(s_{t\!+\!k};\theta_v)\!-\!V(s_t;\theta_v)
\end{equation}

Learning is performed using multiple threads. The environment and agents operate asynchronously in each thread. The parameters $\theta$ and $\theta_v$ are stored in a shared memory. Each thread first copies the shared parameters to a local memory. Each agent uses the local parameters to determine the action, calculates the update amounts, updates the shared parameters, and then copies the updated shared parameters to the local memory again.

\subsection{Example of Reward Functions}
Two reward functions are introduced below as examples of applying RL to a racing simulator. The reward function in \cite{lillicrap2015continuous} is as follows.
\begin{equation}
r_t=\begin{cases}
-1 & (\textrm{collision with wall}) \\
v_t\cos\theta_t & (\textrm{others})
\end{cases}
\end{equation}
Here, $v_t$ is the velocity in the vehicle traveling direction, and $\theta_t$ is the angle relative to the course traveling direction. The reward is process-oriented for velocity, but the penalty for collision is purpose-oriented. Therefore, learning to accelerate is easy, but learning to avoid collision is thought to take time, or easily become unstable. In fact, \cite{perot2017end} reports that despite driving a simple oval course, learning was unstable, with success in some cases and failure in others.

The reward function which is adopted by \cite{perot2017end} is as follows.
\begin{equation}
r_t=v_t(\cos\theta_t-d_t)
\end{equation}
Here, $d_t$ is the offset from the center of the course. This reward is completely process-oriented, and learning is thought to proceed quickly. In addition, it is a simple reward of more quickly traveling straight in the center of the course. However, it is difficult to adjust priorities, such as loosening the restriction to travel in the center and increasing the speed, as well as modifying the function, such as addition of a term related to ride comfort

\section{Design Scheme of Process-Oriented Reward Function}
This chapter proposes a design scheme of a process-oriented reward function that is suited for automated driving. When there are multiple evaluation items, the priority can be adjusted and new evaluation items can also be added.

When designing the reward, first the evaluation items $f_k(s,a,s')$ and evaluation functions $e_k(f)$ are determined.
Here, $k<n$, where $n$ is the number of evaluation items.
The evaluation items must be determined uniquely according to the states and actions.
The evaluation function maps the evaluation items to evaluation values between 0 and 1. The temporary reward $r_{tmp}(s,a,s')$ is the product of all the evaluation values, and is expressed as follows.
\begin{equation}
\label{eq:tmpr}
r_{tmp}(s,a,s')=\prod_{k=0}^{n-1}{e_k(f_k(s,a,s'))}
\end{equation}
The reward $r(s,a,s')$ is given as follows to help prevent state value from decreasing as the state approaches the terminal state.
\begin{equation}
r(s,a,s')=\begin{cases}
r_{tmp}(s,a,s') & (s' \textrm{is not terminal}) \\
\cfrac{r_{tmp}(s,a,s')}{1-\gamma} & (s' \textrm{is terminal})
\label{eq:r}
\end{cases}
\end{equation}

\begin{figure}
\subfigure[Action value when terminal state is not considered]{\includegraphics[clip, width=\columnwidth]{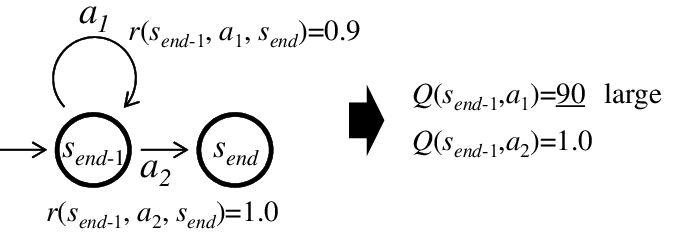}}
\subfigure[Action value when terminal state is considered]{\includegraphics[clip, width=\columnwidth]{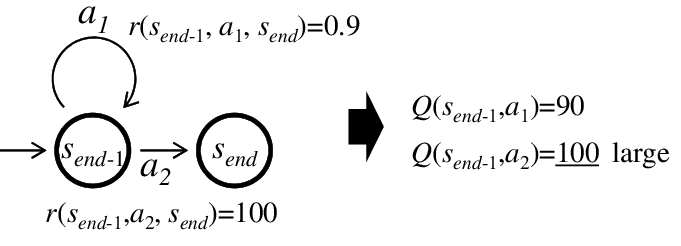}}
\caption{Confirming effect of the equation (\ref{eq:r}) by calculating action values}
\label{fig:state}
\end{figure}
Here, figure \ref{fig:state}(a). shows an example of action value $Q(s,a)$ when the terminal state is not taken into account. The action value is the expected future return value when action $a$ is taken in state $s$. In this example, when action $a_1$ is selected in the state immediately before the terminal state, the same state is maintained and the reward 0.9 is received, and when action $a_2$ is selected the state transitions to the terminal state (goal) and the reward 1.0 is received. The return when $\gamma=0.99$ is expressed by equation (\ref{eq:r}), so when action $a_1$ is continuously selected for infinity, $Q(s_{end-1},a_1) = 0.9/(1-0.99)=90$. The return when action $a_2$ is selected is the same as the final reward, so $Q(s_{end-1},a_2) = 1$. This means that the value of action $a_1$ is higher, even though the reward is higher for action $a_2$, which results in a policy that does not move toward the goal, and instead continues to take action $a_1$. Therefore, the reward when the terminal state is reached is set to the return when a reward of 1.0 is obtained for infinity, that is to say $1.0/(1-0.99)=100$, as shown in figure \ref{fig:state}(b). This increases the value of action $a_2$ and enables to obtain a policy that moves toward the goal.

\begin{figure}
\begin{center}
\includegraphics[width=\columnwidth]{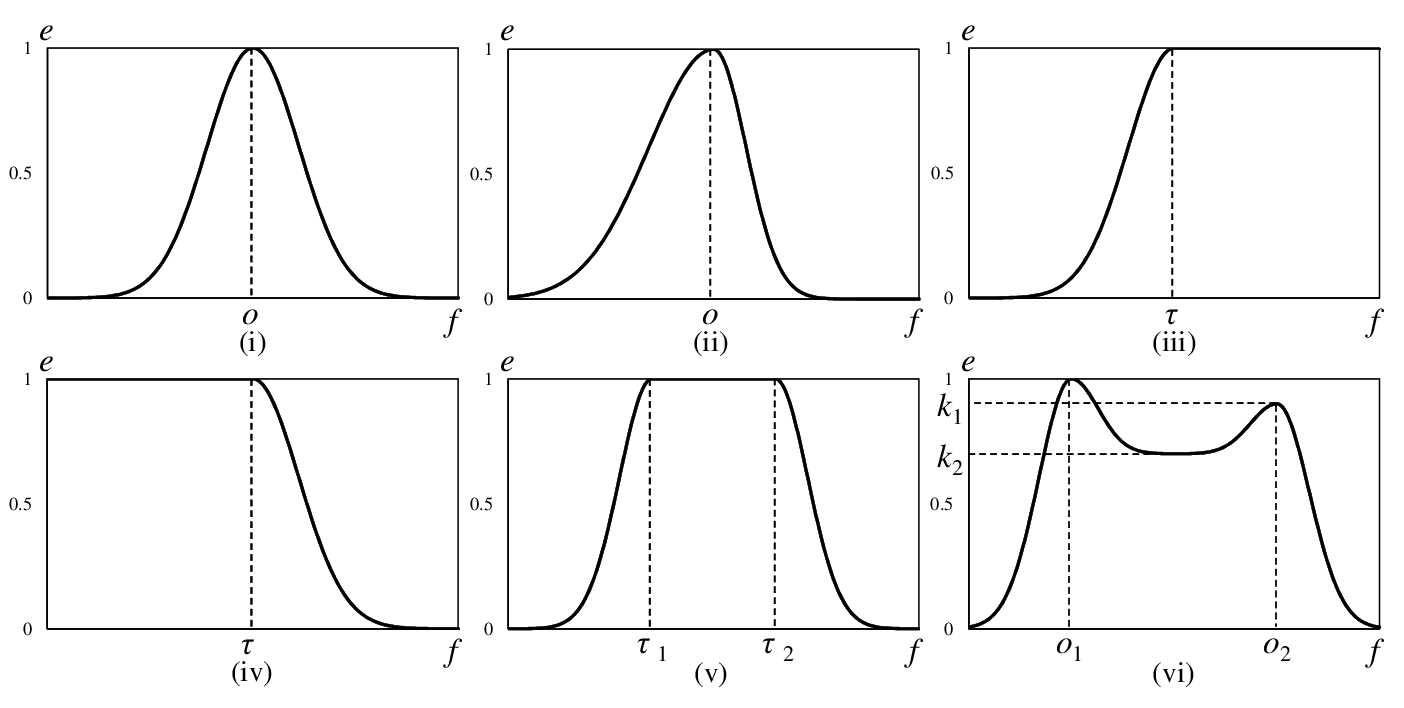}
\caption{Evaluation function for each purpose} 
\label{fig:eval}
\end{center}
\end{figure}

The six different evaluation functions shown in figure \ref{fig:eval}. are proposed in accordance with the demands on the evaluation item $f$. 

\begin{enumerate}
\renewcommand{\labelenumi}{(\roman{enumi})}
\item Evaluation function $f$ has a target value
\end{enumerate}
\begin{equation}
e_k(f)=\exp(-(f-o)^2/\alpha)
\end{equation}
$o$ is the target value, and $\alpha$ is a parameter that adjusts the tolerance for the offset from the target value.

\begin{enumerate}[resume]
\renewcommand{\labelenumi}{(\roman{enumi})}
\item The tolerance changes beyond the target value
\end{enumerate}
\begin{equation}
e_k(f)=\begin{cases}
\exp(-(f-o)^2/\alpha_1) & (f<o)\\
\exp(-(f-o)^2/\alpha_2) & (f{\geq}o)
\end{cases}
\end{equation}
$\alpha_1$ and $\alpha_2$ are parameters that adjust the respective tolerance for offset in the directions below and above the target value. 

\begin{enumerate}[resume]
\renewcommand{\labelenumi}{(\roman{enumi})}
\item Tolerating offset above the threshold value
\end{enumerate}
\begin{equation}
e_k(f)=\begin{cases}
\exp(-(f-\tau)^2/\alpha) & (f<\tau)\\
1 & (f{\geq}\tau)
\label{eq:lower}
\end{cases}
\end{equation}
$\tau$ is the threshold value at which the reward starts to decrease, and $\alpha$ is a parameter that adjusts the tolerance for the offset below the target value. When there is a true lower limit value that cannot be tolerated, set a threshold value that is sufficiently larger than the lower limit value.

\begin{enumerate}[resume]
\renewcommand{\labelenumi}{(\roman{enumi})}
\item Tolerating offset below the threshold value
\end{enumerate}
\begin{equation}
e_k(f)=\begin{cases}
1 & (f<\tau)\\
\exp(-(f-\tau)^2/\alpha) & (f{\geq}\tau)
\label{eq:upper}
\end{cases}
\end{equation}
$\tau$ is the threshold value at which the reward starts to decrease, and $\alpha$ is a parameter that adjusts the tolerance for the offset above the target value. This function is the inverse of equation (\ref{eq:lower}).

\begin{enumerate}[resume]
\renewcommand{\labelenumi}{(\roman{enumi})}
\item There are upper and lower threshold values
\end{enumerate}
\begin{equation}
e_k(f)=\begin{cases}
\exp(-(f-\tau_1)^2/\alpha_1) & (f<\tau_1)\\
1 & (\tau_1{\leq}f<\tau_2)\\
\exp(-(f-\tau_2)^2/\alpha_2) & (f{\geq}\tau_2)
\end{cases}
\end{equation}
$\tau_1$ and $\tau_2$ are the respective upper and lower threshold values at which the reward starts to decrease, and $\alpha_1$ and $\alpha_2$ are parameters that adjust the respective tolerance for offset in the directions below $\tau_1$ and above $\tau_2$. This function is a combination of equation (\ref{eq:lower}) and (\ref{eq:upper}).

\begin{enumerate}[resume]
\renewcommand{\labelenumi}{(\roman{enumi})}
\item There are two target values
\end{enumerate}
\begin{equation}
e_k(f)=\begin{cases}
\exp(-(f-o_1)^2/\alpha_1)\\\hspace{31mm}(f<o_1)\\
(1-k2)\exp(-(f-o_1)^2/\alpha_2)+k2&\\\hspace{31mm}(o_1{\leq}f<\frac{o_1+o_2}{2})\\
(k1-k2)\exp(-(f-o_2)^2/\alpha_3)+k2&\\\hspace{31mm}(\frac{o_1+o_2}{2}{\leq}f<o_2)\\
k_1\exp(-(f-o_2)^2/\alpha_4)&\\\hspace{31mm}(f{\geq}o_2)
\end{cases}
\end{equation}
$o_1$ is the 1st target value and $o_2$ is the 2nd target value. $\alpha_1$, $\alpha_2$, $\alpha_3$, and $\alpha_4$ are the respective tolerances for the offset from the target value in each section. $k_1$ is the 2nd target value priority relative to the 1st target value. $k_2$ is the parameter that determines how easily the state transitions between target values.

In the proposed scheme, the reward is expressed by the product of all evaluation functions, so new evaluation functions can easily be added. In addition, each evaluation function can set the tolerance for the offset from the target value, so the evaluation item priority can be changed. (For example, to give a certain evaluation item priority, reduce the tolerance of that evaluation function.).

\section{Application to Automated Driving}
\subsection{Test Environment}
This chapter describes the application of the proposed scheme to circuit driving and highway cruising to investigate application in regards to automated driving. We use TORCS that is freeware racing simulation with a patch for robot racing and the detail is described in \cite{loiacono2013simulated}. Furthermore, we modify the patch to acquire the positions and relative speeds of ten nearby vehicles within 200m. Table \ref{tb:torcs} shows the inputs from TORCS.

\begin{figure}[tb]
\begin{center}
\includegraphics[width=\columnwidth]{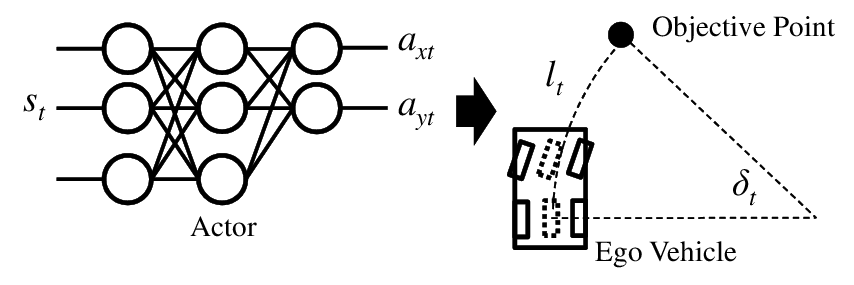}
\end{center}
\caption{Outline of the control scheme using the generated next objective point}
\label{fig:control}
\end{figure}

Figure \ref{fig:control} shows an outline of vehicle control using RL. The actor outputs the acceleration in the local coordinate system, and the controller calculates the target coordinates 0.1s in the future assuming uniform acceleration motion. In addition, the steering angle needed to track the target coordinates is calculated using a 2-wheel model, and the accelerator and brake pedal operations are calculated by PID control. Separation of target coordinate generation and vehicle control enables the controller to absorb differences between vehicle models, so there is no need for relearning depending on the model.

11 courses are used for circuit driving. Learning was performed asynchronously on 10 of the 11 courses to increase the generalization performance. In addition, another course was not used for learning, and was only used as the test course. Circuit driving assumed the condition without other vehicles.

An oval course with a total length of 1770m and models of other vehicles were created to simulate highway cruising. The course has two lanes, a traveling lane and a passing lane, and the other vehicles continuously cruise in the same lane at a velocity of 90 km/h in the traveling lane and 130 km/h in the passing lane.

\begin{table}[tb]
\begin{center}
\caption{Input from the modded TORCS}\label{tb:torcs}
\vspace{-3mm}
\begin{tabular}{cl}
Items &Description\\\hline 
Angle        &0 is along the track.\\
Velocity &$v_x$ and $v_y$ in the local coord.\\
Track position   &Distance from the center of the track.\\
Track  &Distance to edge of the track(Each 20 degree).\\
Others' positon &Near 10 cars' (x, y) in local coord.\\
Others' speed &Near 10 cars' relative speed.\\\hline
\vspace{2mm}
\end{tabular}
\end{center}
\end{table}

\subsection{Reward Function}
\begin{figure}[t]
\begin{center}
\includegraphics[width=70mm]{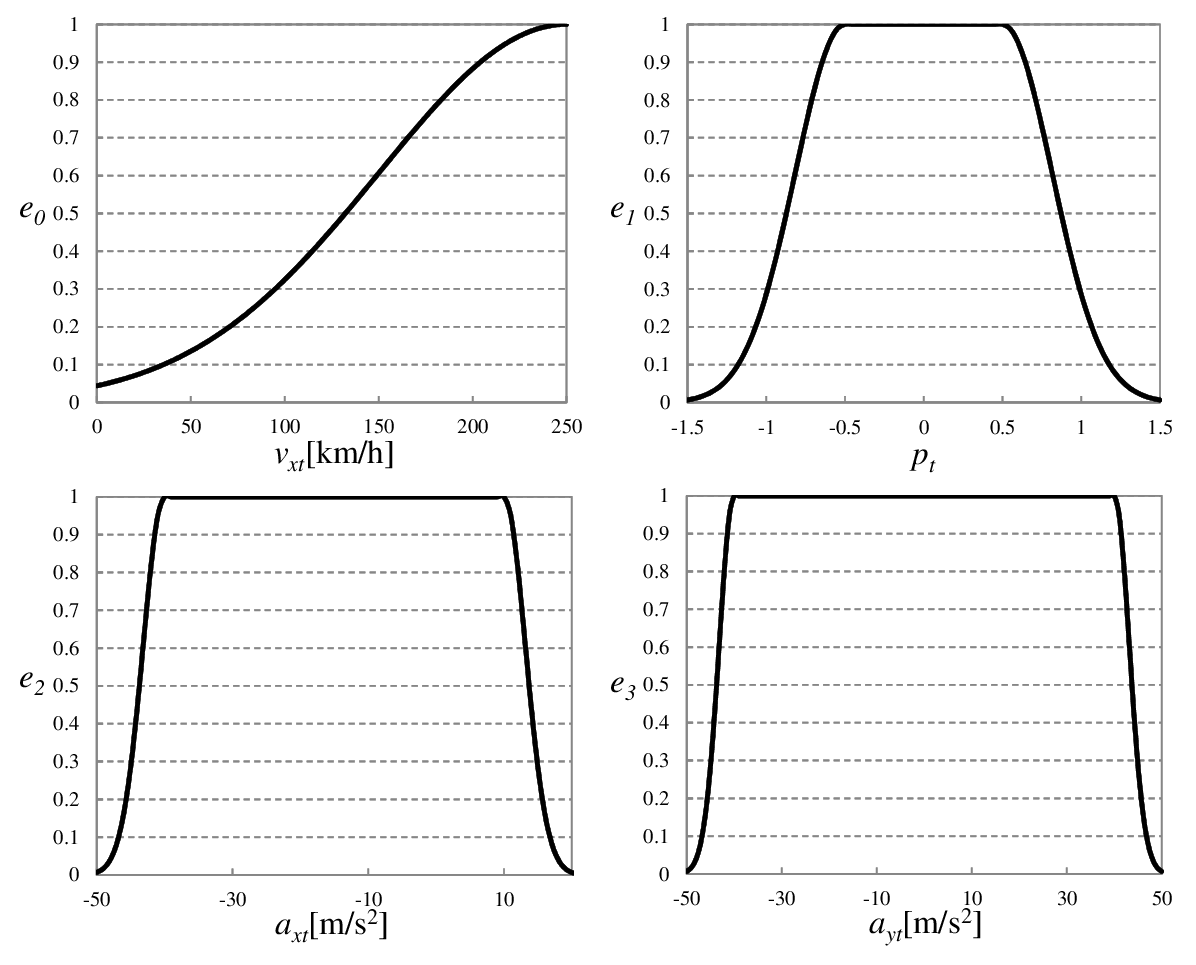}
\end{center}
\caption{Evaluation functions for circuit driving}
\label{fig:eval_circuit}
\end{figure}

The evaluation functions for circuit driving are as follows (figure \ref{fig:eval_circuit}).
\begin{eqnarray}
e_0&=&\exp(-(v_{xt}-250)^2/20000) \nonumber \\
e_1&=&\begin{cases}
\exp(-(p_t+0.5)^2/0.2) & (p_t<-0.5)\\
1 & (-0.5{\leq}f_1<0.5)\\
\exp(-(p_t-0.5)^2/0.2) & (p_t{\geq}0.5)
\end{cases} \nonumber \\
e_2&=&\begin{cases}
\exp(-(a_{xt}+30)^2/20) & (a_{xt}<-30)\\
1 & (-30{\leq}f_2<10)\\
\exp(-(a_{xt}+10)^2/20) & (a_{xt}{\geq}10)
\end{cases} \\
e_3&=&\begin{cases}
\exp(-(a_{yt}+40)^2/20) & (a_{yt}<-40)\\
1 & (-40{\leq}a_{yt}<40)\\
\exp(-(a_{yt}-40)^2/20) & (a_{yt}{\geq}40)
\end{cases} \nonumber
\end{eqnarray}
Here, $v_{xt}$ is the velocity [km/h] in the vehicle traveling direction, $p_t$ is the distance from the center of the course (-1 is the left edge and +1 is the right edge), and $a_{xt}$ and $a_{yt}$ are the respective acceleration [m/s2] in the traveling direction and the lateral direction output by the actor. Therefore, the reward is higher when the velocity is close to 250 km/h, the vehicle does not deviate too far from the center of the track, and the output acceleration is not too large.

\begin{figure}[tb]
\begin{center}
\includegraphics[width=70mm]{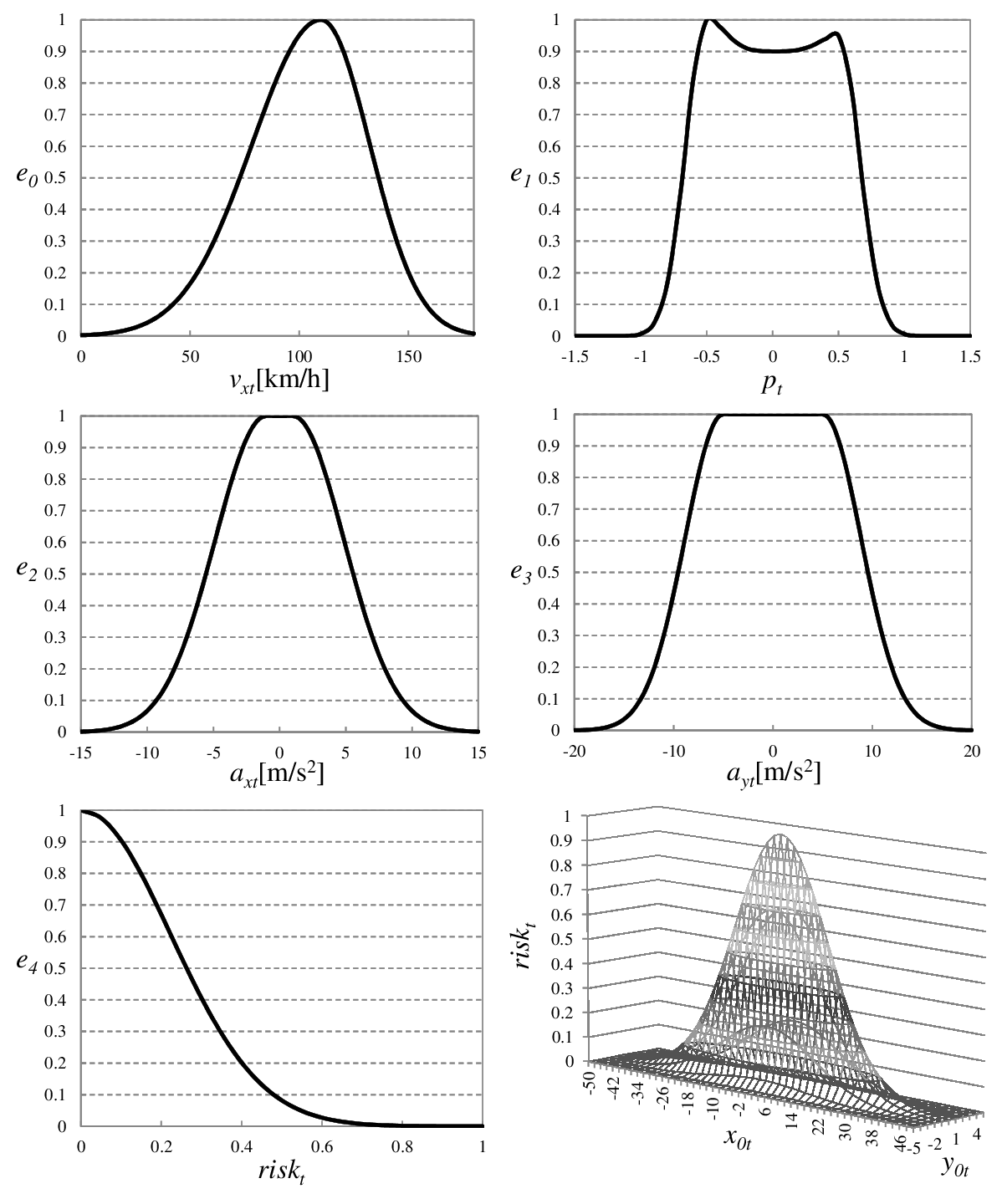}
\end{center}
\caption{Evaluation functions for highway cruising}
\label{fig:eval_highway}
\end{figure}

For highway cruising, an evaluation item as follows is added for avoiding risk with other cars.
\begin{equation}
risk_t=\sum_{k=0}^{n-1}\exp(x_{kt}^2/400+y_{kt}^2/3)
\end{equation}
The evaluation functions is as follows (figure \ref{fig:eval_highway}).
\begin{eqnarray}
e_0&=&\begin{cases}
\exp(-(v_{xt}-110)^2/2000) & (v_{xt}<110)\\ 
\exp(-(v_{xt}-110)^2/1000) & (v_{xt}{\geq}110)
\end{cases} \nonumber \\
e_1&=&\begin{cases}
\exp(-(p_t+0.5)^2/0.05) & (p_t<-0.5)\\
0.1\exp(-(p_t0.5)^2/0.04)+0.9 & (-0.5{\leq}p_t<0)\\
0.05\exp(-(p_t-0.5)^2/0.04)+0.9 & (0{\leq}p_t<0.5)\\
0.95\exp(-(p_t-0.5)^2/0.05) & (p_t{\geq}0.5)
\end{cases}  \nonumber \\
e_2&=&\begin{cases}
\exp(-(a_{xt}+1)^2/30) & (a_{xt}<-1)\\
1 & (-{\leq}f_2<1)\\
\exp(-(a_{xt}-1)^2/30) & (a_{xt}{\geq}1)
\end{cases} \\
e_3&=&\begin{cases}
\exp(-(a_{yt}+5)^2/30) & (a_{yt}<-5)\\
1 & (-5{\leq}f_3<5)\\
\exp(-(a_{yt}-5)^2/30) & (a_{yt}{\geq}5)
\end{cases} \nonumber \\
e_4&=&\exp(-(risk_t)^2/0.1)  \nonumber
\end{eqnarray}
Figure \ref{fig:eval_highway} also shows the evaluation functions and the $risk_t$ for the case when there is one other vehicle. The reward increases when the velocity is close to 110 km/h, the output acceleration is not too large, and the distance from the other vehicle increases. When there is no vehicle in front, the highest reward is obtained by driving in the left lane at a constant velocity of 110 km/h. However, when driving continuously at 110 km/h, the ego vehicle catches up to the other vehicle driving at 90 km/h. At this time, the reward is higher when the ego vehicle drives at 110 km/h in the right lane than when decelerating to 90 km/h in the left lane.

\subsection{Learning Results of Circuit Driving}
\begin{figure}[tb]
\begin{center}
\includegraphics[width=\columnwidth]{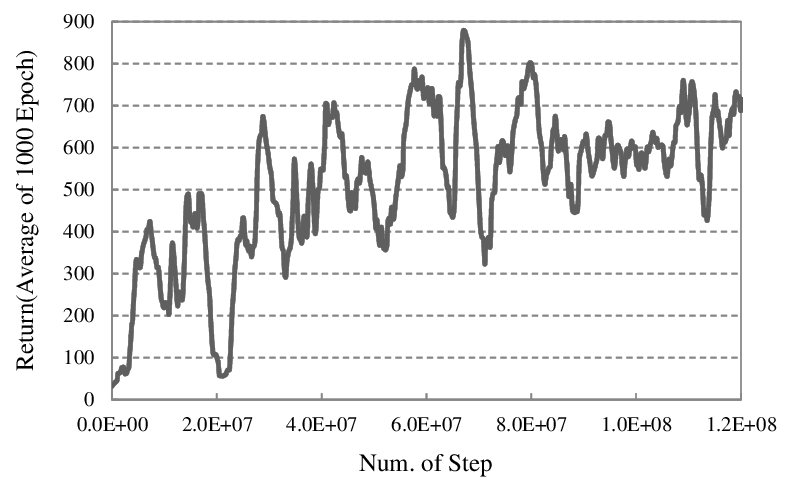}
\end{center}
\caption{Return during training in circuit driving}
\label{fig:return_circuit}
\end{figure}

Figure \ref{fig:return_circuit} shows the average return for each 1000 epochs in order to confirm learning convergence. Convergence was judged and learning ended at $1.2\times10^8$ steps. This number of steps corresponds to 139 days of time in the simulation.

\begin{figure}[tb]
\begin{center}
\includegraphics[width=\columnwidth]{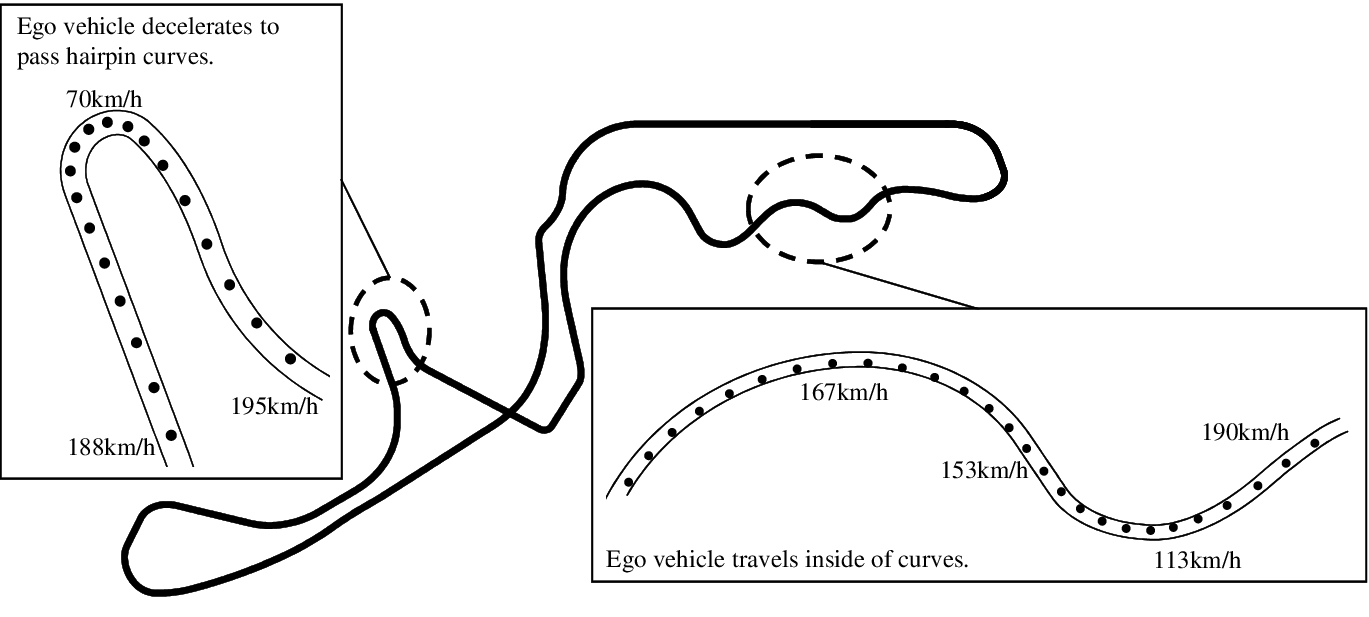}
\end{center}
\caption{Result of trajectory when driving circuit}
\label{fig:circuit}
\end{figure}

Figure \ref{fig:circuit} shows the trajectory (0.5s interval) through an S-shaped curve and hairpin curve when driving on the test course using the obtained policy. This shows that an appropriate policy could be obtained, such as decelerating before the curves and driving the shortest distance through the S-shaped curve.

\subsection{Learning Results for Highway Cruising}
\begin{figure}[tb]
\begin{center}
\includegraphics[width=\columnwidth]{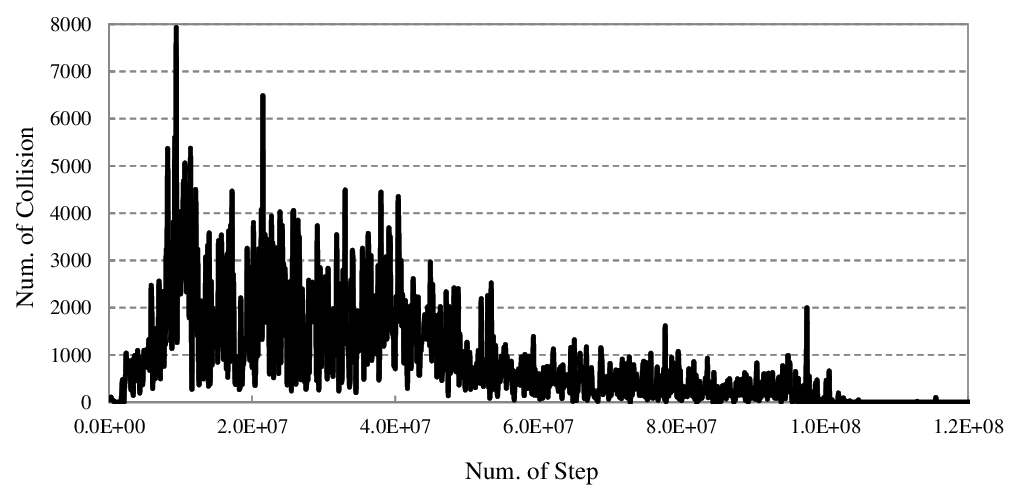}
\end{center}
\caption{Collision times during training in highway cruising}
\label{fig:collision_highway}
\end{figure}

Figure \ref{fig:collision_highway} shows the number of collisions per 1000s of game time in order to confirm learning convergence. Convergence was judged and learning ended at $1.2\times10^8$ steps same as the previous experiment.

\begin{figure}[tb]
\begin{center}
\includegraphics[width=\columnwidth]{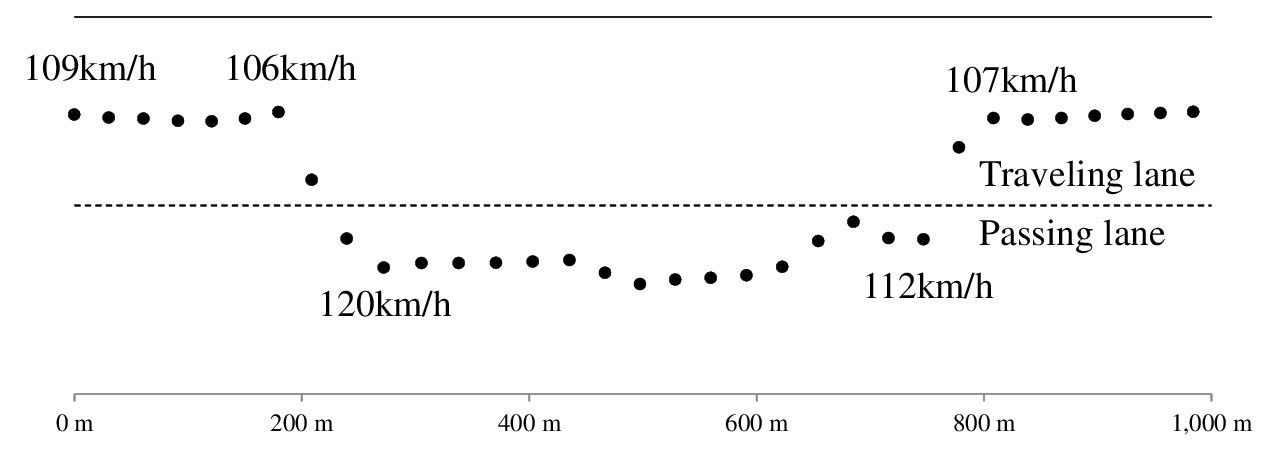}
\end{center}
\caption{Result of trajectory when overtaking}
\label{fig:overtake}
\end{figure}

Next, Figure \ref{fig:overtake} shows the passing track (1s interval) when driving with the obtained policy. Lane change started when the ego vehicle closed to approximately 30m from the vehicle in front. The reason of the deceleration in the left lane is thought to be avoiding catching up to the front vehicle. Conversely, the reason for acceleration to 110km/h or more in the right lane is that a rear vehicle driving at 130km/h does not catch up to the ego vehicle. The vehicle behavior in the passing lane exhibits some unstable motion. The reason for this is thought to be that the reward between the traveling lane and the passing lane in evaluation function $e_1$ is too high. Therefore, smoother driving might be obtained by enhancing the evaluation function parameters.

\section{Conclusion and Future Work}
This paper proposed a reward function design scheme for automated driving. The proposed reward function evaluates driving states and actions on the process to the destination. Moreover, the scheme can add multiple evaluation items to the reward function. In automated driving, various items should be evaluated, such as velocity, passengers’ comfort, distance from obstacles, and so on. Therefore, the scheme is suited for automated driving.

The proposed scheme was also applied to circuit driving and highway cruising to investigate application to automated driving. The results showed that in circuit driving, appropriate driving positions were obtained, such as traveling on the inside of corners, and rapid deceleration to turn along sharp curves was also learned. In highway cruising, lane change in an environment where other vehicles are present, deceleration to avoid catching up to a vehicle in front, and acceleration so that a vehicle to the rear does not catch up to the ego vehicle were learned.

As improving algorithm, risk sensitive reinforcement learning developed by \cite{mihatsch2002risk} will be introduced for more safety.  In addition, it is also important issue to adjust parameters in the reward function. We are planning to apply inverse reinforcement learning(IRL) cf. \cite{ng2000algorithms, abbeel2010autonomous} that can estimate reward function using expert action data. We consider that the reward function which is related to safety should be hand-crafted, but the one related to comfortability can be estimated automatically with IRL by \cite{kishikawa2019comfortable}. Therefore, we will develop the scheme combining the hand-crafted reward and the generated reward by IRL.

\bibliography{ifacconf}

\appendix
\section{Learning Parameters}    
Table \ref{tb:parameter} shows the neural network configuration and learning parameters. The neural network has four fully linked layers consisting of an input layer, two intermediate layers, and an output layer. A Rectified Linear Unit (ReLU) function is inserted between the input layer and intermediate layers, and between the intermediate layers. The critic output layer outputs the product-sum operation results as is. The actor output layer outputs the mean and variance for each action. The mean side inserts tan-1 and outputs a value between -1 and 1. The variance side inserts a softmax function and outputs a value of 0 or more. In addition, an offset is added to the variance to promote exploration.

\begin{table}[h]
\caption{Learning parameters}
\label{tb:parameter}
\begin{center}
\begin{tabular}{ll}
Items & Values \\\hline
Hidden layer        & $500\times500$ \\
Variance Offset   & 0.04 \\
Discount Rate &0.99 \\
$t_{max}$ & 40 \\
Optimizer  & SharedRMSProp \\
Learning Rate(Actor) & 0.0001 \\
Learning Rate(Critic) & 0.00005 \\\hline
\end{tabular}
\end{center}
\end{table}

\end{document}